\documentclass{article}
\usepackage{spconf}
\usepackage{graphicx}
\usepackage{setspace}
\usepackage{amsmath}
\usepackage{dsfont}
\usepackage{amsfonts}
\usepackage{lipsum} 
\usepackage{multicol}
\usepackage{float}
\usepackage{bm}
\usepackage{color}
\usepackage{etoolbox}
\usepackage{soul}
\usepackage{amssymb}
\usepackage[linesnumbered,ruled,vlined]{algorithm2e}
\usepackage{mathtools,amssymb}
\usepackage{caption}
\usepackage{float}
\usepackage{amsthm}
\usepackage[caption = false,subrefformat=parens,labelformat=parens]{subfig}
\usepackage{adjustbox}
\usepackage{afterpage}
\usepackage{multirow}
\usepackage{mathrsfs}
\usepackage{booktabs}
\usepackage{url}

\title{Fast-Convergent Federated Learning via Cyclic Aggregation}

\name{Youngjoon Lee, Sangwoo Park${}^\dagger$, Joonhyuk Kang
\thanks{This research was supported by the MSIT (Ministry of Science and ICT), Korea, under the ITRC (Information Technology Research Center) support program (IITP-2020-0-01787) supervised by the IITP (Institute of Information \& Communications Technology Planning \& Evaluation)}}
\address{KAIST, South Korea \\
${}^\dagger$King’s College London, United Kingdom
}

\begin{document}

\maketitle

\begin{abstract}
Federated learning (FL) aims at optimizing a shared global model over multiple edge devices without transmitting (private) data to the central server. While it is theoretically well-known that FL yields an optimal model -- centrally trained model assuming availability of all the edge device data at the central server --  under mild condition, in practice, it often requires massive amount of iterations until convergence, especially under presence of statistical/computational heterogeneity. This paper utilizes \emph{cyclic learning rate} at the server side to reduce the number of training iterations with increased performance without any additional computational costs for both the server and the edge devices. Numerical results validate that, simply plugging-in the proposed cyclic aggregation to the existing FL algorithms effectively reduces the number of training iterations with improved performance.
\end{abstract}
\begin{keywords}
machine learning, distributed learning, federated learning, edge computing, privacy-preserved, cyclic learning rate
\end{keywords}
\section{Introduction}
\label{sec:intro}
Given sufficient data from multiple sources, deep learning \cite{goodfellow2016deep} outperforms conventional model-based approaches as it can approximate any underlying function with deep neural networks via data-driven approach \cite{simeone2022machine}. While the strength of deep learning comes at the cost of massive amount of data \cite{simeone2022machine}, in many real-world scenarios, such data are often not shared to the central server, due to privacy issues \cite{niknam2020federated,khan2021federated}. Federated learning (FL) \cite{mcmahan2017communication} mitigates this problem via asking for updated \emph{models} at the edge device side instead of their data sets. By aggregating locally updated models obtained from the edge devices, FL achieves the same performance to that of central learning which assumes availability of the whole local devices' data \cite{li2019convergence}.

However, in practice, FL requires sufficient amount of communication rounds between the central server and the edge devices in order to reach the desired performance level \cite{zhou2018convergence}. 
This becomes more severe when the edge devices' data are generated in a non-iid manner \cite{zhao2018federated}, and/or when the computational power varies across different edge devices \cite{li2020federated}. Such statistical/computational heterogeneity often \emph{drifts} the local models towards their local minima to harm the aggregation performance at the server side that generally degrades the performance \cite{kairouz2021advances} and increase the number of communication rounds to converge \cite{krizhevsky2009learning}.
\begin{figure}
     \centering
     \includegraphics[width=\columnwidth]{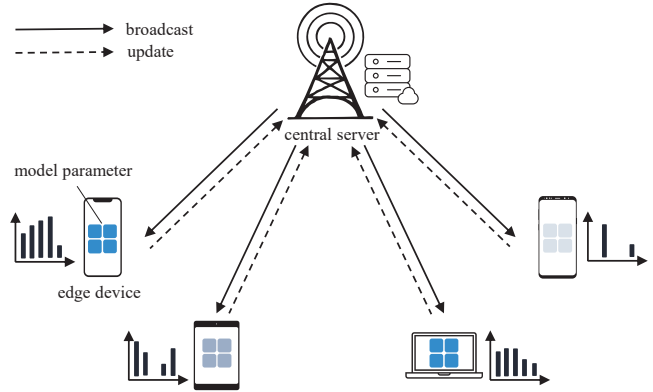}
     
     \caption{Illustration of FL in statistical/computational heterogeneous environment: Statistical heterogeneity comes from non i.i.d. data samples for each edge device, and computational heterogeneity stems from different computing power per device.}
     \label{fig:1}
     \vspace{-0.5cm}
\end{figure}

Recently, several studies have alleviate such \emph{client-drift} phenomenon by improving the local client training process \cite{li2018federated, li2021model, li2021edrs}.
By considering proximal regularization \cite{li2018federated}, adding contrastive loss term \cite{li2021model}, or applying restricted-softmax function \cite{li2021edrs} during local training, client-drift has been successfully alleviated at the cost of increased complexity at the client side.
In this paper, we mitigate client-drift in heterogeneous FL \emph{without} any additional cost either at the client or the server side. Key idea is to improve the generalization performance of the aggregated model at the server side by applying cyclic learning rate \cite{smith2017cyclical} that varies across communication rounds.

From the initial proposal of cyclic learning rate \cite{smith2017cyclical}  that changes the learning rate periodically during training iterations, it has been widely applied as a means to achieve improved performance with reduced learning time \cite{smith2017exploring}, \cite{smith2022general}, yet general from of proof is still an open problem \cite{goujaud2022super}. To the best of authors' knowledge, this paper is first to apply cyclic learning rate during aggregation at the server side for FL to speed up FL process in heterogeneous environment.

Main contributions of this paper are as follows:
\begin{itemize}
    \item We propose cyclic learning rate during aggregation at the server side, referred to as \emph{cyclic aggregation}, that can be applied to most of the existing FL schemes. 
    \item Cyclic aggregation enables fast convergent and increased performance without additional computational costs.
    \item By applying cyclic aggregation to state-of-the-art FL schemes, e.g., FedAvg \cite{mcmahan2017communication}, FedProx \cite{li2018federated}, MOON \cite{li2021model}, and FedRS \cite{li2021edrs}, we numerically show that the proposed approach effectively reduces the training time with increased performance on MNIST \cite{deng2012mnist}, FMNIST \cite{xiao2017fashion}, CIFAR-10 \cite{krizhevsky2009learning}, and SVHN \cite{netzer2011reading} dataset in the presense of \emph{label skew} \cite{li2021edrs}.
\end{itemize}

\section{Model and Problem}\label{sec:system}
\begin{figure}[h]
     \centering
     \includegraphics[width=\columnwidth]{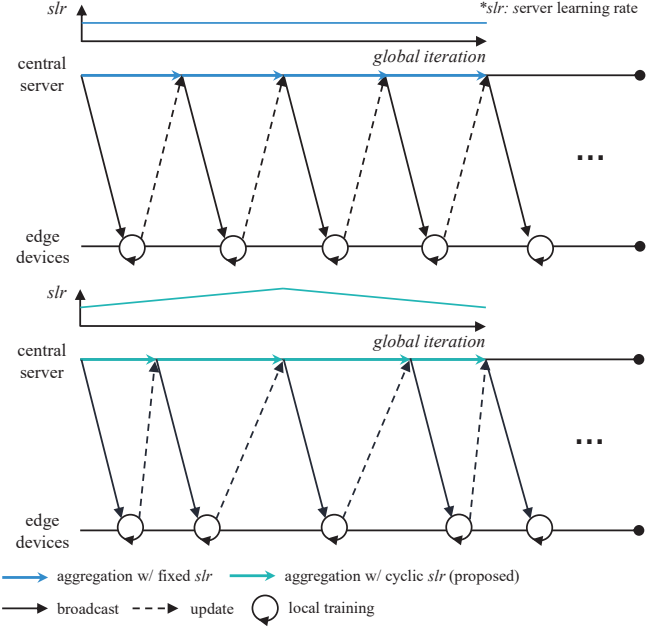}
     
     \caption{Comparison between the cyclic aggregation (Top) and fixed aggregation (Bottom) during FL at each . Solid black arrows denote ``broadcast'' and black dotted line denote ``update''. ``local training'' is described by circled arrow. Solid green and blue arrows denote the cyclic and fixed aggregation respectively.}
     \label{fig:2}
     \vspace{-0.5cm}
\end{figure}

The proposed FL framework aided by cyclic learning rate \cite{smith2017cyclical} consists of the following steps, which are repeated until convergence (see Fig. \ref{fig:2}):
\begin{itemize}
    \item \textbf{Broadcast:} The central server samples $M$ edge devices randomly and transmit the global model to all edge devices.
    \item \textbf{Update:} Randomly selected edge devices update the shared global model and transmit the locally updated model to the central server.
    \item \textbf{Cyclic Aggregation:} The central server aggregate the transmitted models by multiplying cyclic server learning rate.
\end{itemize}
Note that conventional FL treats the server learning rate to be a fixed value \cite{kairouz2021advances}.

\subsection{Federated heterogeneous setting}
In this section, we describe mathematical objective of FL and introduce standard FL technique, FedAvg \cite{mcmahan2017communication}.
As depicted in Fig. \ref{fig:1}, we assume $N$ edge devices constitute the federated heterogeneous network, communicating through the central server.
Each edge device $n=1,\ldots,N$ holds private dataset $\mathcal{D}^{n}$ with different number of data points.
Furthermore, the edge devices have a wide variety of computational capabilities, which affects the number of iterations for local update.
The standard goal of FL is to optimize a shared global model $\theta \in \Theta$ among model parameter space $\Theta$ without sharing the entire edge devices' dataset $\mathcal{D}=\cup_{n=1}^{N}\mathcal{D}^{n}$ to the central server. Accordingly, the training objective of FL is to jointly solve the optimization problem $\min_{\theta \in \Theta}F(\theta) \triangleq \frac{1}{N} \sum_{n=1}^{N}F_{n}(\theta)$, where $F(\theta)$ is the global objective function at the central server, and the local objective function for the $n^{th}$ edge device $F_n(\theta)$ is defined as $F_{n}(\theta)=\frac{1}{|\mathcal{D}^{n}|}\sum_{x \in \mathcal{D}^{n}}f_{n}(\theta; x),$ where $f_{n}(\theta; x)$ denotes loss function of the learning model and $x$ represents the data sampled from the local dataset ${D}^{n}$.
For $i \neq j$, the data distribution of $D^i$ and $D^j$ may be quite different.

\subsection{Concept of cyclic server learning rate $\gamma_{cyclic}$}
The cyclic server learning rate concept is based on \cite{smith2017cyclical} that allows increase and decrease in the learning rate, which may bring some negative impact in short-term wise, but often results in a long-term positive benefit in centralized learning \cite{smith2022general}. To this end, we propose to adjust server learning rate as
\begin{align}
\label{eq:cyclic_slr}
\gamma_{cyclic}(i_g) = \gamma_{fixed} - a\bigg(\frac{1}{2} - \frac{1}{\pi}\sum_{n=1}^{\infty} (-1)^{n}\frac{\sin(2 \pi k f i_g)}{k}\bigg),
\end{align}
with predetermined amplitude $a$ and frequency $f$ that oscillates around fixed learning rate $\gamma_{fixed}$.

\begin{figure}[h]
     \centering
     \includegraphics[width=\columnwidth]{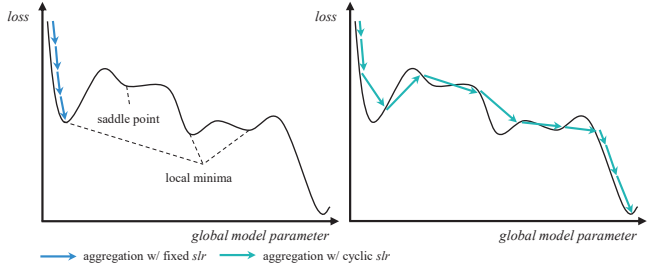}
    
     \caption{Illustration of learning process with fixed learning rate $\gamma_{fixed}$ and cyclic learning rate $\gamma_{cyclic}$. Fixed learning rate $\gamma_{fixed}$ has a constant value (length of the arrow), while cyclic learning rate $\gamma_{cyclic}$ changes its value cyclically for every iteration $i_g$.}
     \label{fig:3}
     \vspace{-0.7cm}
\end{figure}

We now provide intuitive illustration why cyclic server learning rate \eqref{eq:cyclic_slr} improves generalization performance with improved convergence speed. As mentioned in Sec.~\ref{sec:intro}, mathematical proof of cyclic aggregation for FL is out of scope of this paper.
In Fig.~\ref{fig:3}, we depict some loss landscape that corresponds to $F(\theta)$. According to \cite{dauphin2015equilibrated}, the difficulty of gradient-based optimization lies in the presence of saddle points. Compared to fixed learning rate (left of Fig. 3), cyclic learning rate (right of Fig. 3) enables to traverse local minima and saddle points \cite{smith2017cyclical} via dynamic gradient updates.
\subsection{Fast-convergent FL}
In FL, the central server communicates with the edge devices for $G$ global iterations to minimize $F(\theta)$.
In this section, we describe the process of the proposed fast-convergent FL during one global iteration $i_g$.
At first the central server randomly samples a subset $M \ll N$ of total edge devices.
Then the central server broadcast its aggregated model $\theta_{i_g}$ to each edge device.
Note that the global model is randomly initialized at the very first global iteration $i_g=0$.

After receiving $\theta_{i_{g}}$ from the central server, all edge devices initialize it to its model as $\theta_{i_g}^{n} \leftarrow \theta_{i_g},$ where $\theta_{i_{g}-1}^{n}$ is the local model of each $n^{th}$ edge device.
Then randomly selected $M$ edge devices starts the update process $EdgeOpt$ according to its FL technique in parallel. 
For instance, FedAvg considers SGD \cite{simeone2022machine} as $EdgeOpt$.
Precisely, randomly selected $m \in M$ edge devices update its local model with $D^{m}$ up to the maximum number of local epochs $i_l=L^{m}$ as follows:
\begin{align}\label{relativity2}
\theta_{i_g}^{m,i_l=L^{m}} &\leftarrow EdgeOpt(\theta_{i_g}, D^{m}, L^{m}).
\end{align}
Once the local training is complete, each selected edge device sends its locally updated model $\theta_{i_g}^{m,i_l=L^{m}}$ to the central server.

After receiving set of models $\{ \theta_{i_g}^m \}_{m=1}^M$, the central server cyclically aggregates these locally updated $M$ models by multiplying $\gamma_{cyclic}$ with FL technique at the server-side as
\begin{align}
\theta_{i_g+1} \leftarrow \gamma_{cyclic}(i_g) \cdot ServerOpt(\{ \theta_{i_g}^m \}_{m=1}^M),
\end{align}
where $ServerOpt(\cdot)$ can be any existing optimization-based FL method at the server-side. 
For instance, FedAvg considers traditional averaging as $ServerOpt$.
After all, $\theta_{i_{g}+1}$ is used for the next global iteration $i_g$ and repeat the procedure until $i_g$ reaches predefined sufficient value $G$. Overall procedure is described in Algorithm \ref{alg_1}.

\begin{algorithm}
\DontPrintSemicolon
\textbf{Server execution:}\\
    \For {global iteration $i_g$=0,...,$G$}
    {
        \textbf{Broadcast:} \\
        Sample $M$ edge devices \\
        Central server transmits $\theta_{i_g}$ to all $N$ edge devices \\
        \textbf{Update:} \\
        \For {edge device $m$=1,...,$M$ \textbf{\textup{in parallel}}}
        {
            
            $\theta_{i_g}^{m,i_l=L^{m}} \leftarrow EdgeOpt(\theta_{i_g}, D^{m}, L^{m})$ \\
        }
        \textbf{Aggregation:}\\
        $\theta_{i_{g}+1} \leftarrow \gamma_{cyclic}(i_g) \cdot ServerOpt(\{ \theta_{i_g}^m \}_{m=1}^M)$ \\
    }
\caption{Fast-convergent FL}
\label{alg_1}
\end{algorithm}

\begin{figure*}
     \centering
     \includegraphics[width=\textwidth]{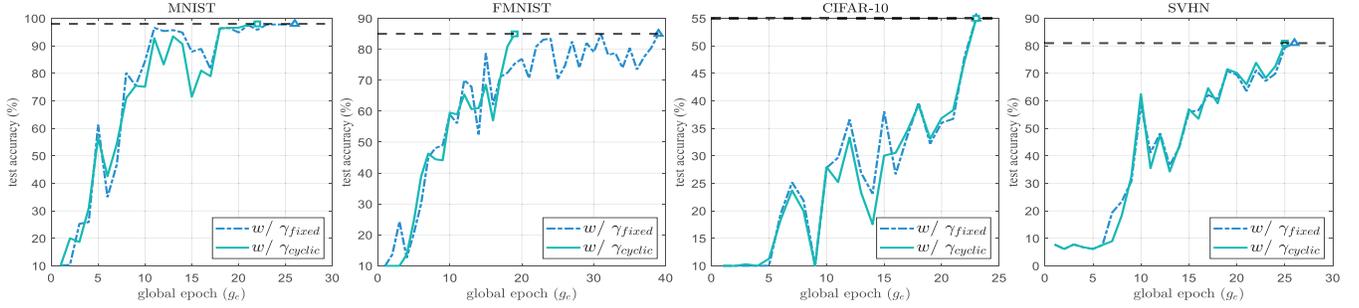}
     
     \caption{Test accuracy with respect to $i_g$ for federated image classification tasks. Solid green line and blue dotted line denote FedAvg with $\gamma_{cyclic}$ and $\gamma_{fixed}$ respectively.}
     \label{fig:4}
     \vspace{-0.1cm}
\end{figure*}

\begin{table*}[!ht]
\centering
\caption{Convergence results on various image classification tasks. The columns denote the minimum number of $i_g$ required to reach the target performance $T$ with state-of-the-art FL techniques.}
\large
\resizebox{\textwidth}{!}{%
\begin{tabular}{@{}lcccccccccccc@{}}
\toprule
\multirow{2}{*}{\textbf{Method}} &
  \multicolumn{3}{c}{MNIST, $T=98\%$} &
  \multicolumn{3}{c}{FMNIST, $T=85\%$} &
  \multicolumn{3}{c}{CIFAR-10, $T=55\%$} &
  \multicolumn{3}{c}{SVHN, $T=81\%$} \\
  \cmidrule(lr){2-4} \cmidrule(lr){5-7} \cmidrule(lr){8-10} \cmidrule(lr){11-13}
 &
  w/ $\gamma_{fixed}$ &
  w/ $\gamma_{cyclic}$ &
  speedup &
  w/ $\gamma_{fixed}$ &
  w/ $\gamma_{cyclic}$ &
  speedup &
  w/ $\gamma_{fixed}$ &
  w/ $\gamma_{cyclic}$ &
  speedup &
  w/ $\gamma_{fixed}$ &
  w/ $\gamma_{cyclic}$ &
  speedup \\ \midrule
FedAvg  & 25 & 21 & $\times$1.19 & 38 & 18 & $\times$2.11 & 40 & 22 & $\times$1.81 & 25 & 24 & $\times$1.04 \\
FedProx & 25 & 13 & $\times$1.92 & 50 & 27 & $\times$1.85 & 42 & 22 & $\times$1.91 & 25 & 24 & $\times$1.04 \\
MOON & 34 & 22 & $\times$1.54 & 31 & 23 & $\times$1.35 & 42 & 22 & $\times$1.91 & 30 & 23 & $\times$1.30 \\
FedRS & 20 & 19 & $\times$1.05 & 32 & 25 & $\times$1.28 & 35 & 33 & $\times$1.06 & 35 & 27 & $\times$1.30 \\ \bottomrule
\end{tabular}%
}
\label{table:1}
\vspace{-0.3cm}
\end{table*}

\begin{table*}[!ht]
\centering
\caption{Performance results on various on image classification tasks. The columns denote the maximum test accuracy over sufficient $G$ with state-of-the-art FL techniques. Six independent experiments are considered to obtain the averaged performance.}
\scriptsize
\resizebox{\textwidth}{!}{%
\begin{tabular}{@{}lcccccccc@{}}
\toprule[0.5pt]
\multirow{2}{*}{\textbf{Method}} &
  \multicolumn{2}{c}{MNIST, $G=200$} &
  \multicolumn{2}{c}{FMNIST, $G=200$} &
  \multicolumn{2}{c}{CIFAR-10, $G=200$} &
  \multicolumn{2}{c}{SVHN, $G=200$} \\ 
  \cmidrule(lr){2-3} \cmidrule(lr){4-5} \cmidrule(lr){6-7} \cmidrule(lr){8-9}
 &
  w/ $\gamma_{fixed}$ &
  w/ $\gamma_{cyclic}$ &
  w/ $\gamma_{fixed}$ &
  w/ $\gamma_{cyclic}$ &
  w/ $\gamma_{fixed}$ &
  w/ $\gamma_{cyclic}$ &
  w/ $\gamma_{fixed}$ &
  w/ $\gamma_{cyclic}$ \\ 
  \midrule
FedAvg & 98.55$\pm$0.18 & 98.61$\pm0.16$ & 84.89$\pm$1.67 & 85.47$\pm$0.97 & 54.52$\pm$3.15 & 55.52$\pm$1.52 & 81.01$\pm$1.29 & 82.32$\pm$0.73 \\
FedProx & 98.61$\pm$0.23 & 98.63$\pm0.18$ & 85.01$\pm$1.23 & 85.52$\pm$1.18 & 56.67$\pm$3.32 & 56.95$\pm$1.71 & 81.61$\pm$0.86 & 82.49$\pm$0.41 \\
MOON  & 98.64$\pm$0.26 & 98.71$\pm$0.15 & 85.65$\pm$1.18 & 86.63$\pm$1.03 & 57.91$\pm$3.28 & 60.94$\pm$1.47 & 82.13$\pm$2.82 & 83.93$\pm$0.61 \\
FedRS & 98.71$\pm$0.08 & 98.73$\pm$0.19 & 86.72$\pm$1.57 & 88.17$\pm$0.62 & 62.16$\pm$1.79 & 63.15$\pm$0.52 & 85.46$\pm$0.66 & 88.68$\pm$0.63 \\
\bottomrule[0.5pt]
\end{tabular}
}
\label{table:2}
\vspace{-0.3cm}
\end{table*}

\section{Experiment and Results}\label{sec:experiment}
\subsection{Experiment setting}
In order to validate the performance of the proposed cyclic aggregation, we consider FL for image classification, i.e,  MNIST \cite{deng2012mnist}, FMNIST \cite{xiao2017fashion}, CIFAR-10 \cite{krizhevsky2009learning}, and SVHN \cite{netzer2011reading} dataset with ResNet-18 \cite{he2016deep} classifier. We apply cyclic learning rate $\gamma_{cyclic}$ to the existing FL algorithms FedAvg \cite{mcmahan2017communication}, FedProx \cite{li2018federated}, MOON \cite{li2021model}, and FedRS \cite{li2021edrs}, which are commonly used FL schemes especially for heterogeneous environments.
We adopt label skew \cite{li2021edrs} to bring statistical heterogeneity by distributing the non-i.i.d. data over $N=100$ edge devices: Each edge device has a maximum of $100$ data for each class, and there are a total of $4$ classes. Computational heterogeneity is considered by randomly selecting $L^m$ between 1 and 5 for every global iteration $i_g$. For the amplitude $a$ and frequency $f$ in \eqref{eq:cyclic_slr}, we perform grid search of $0.1\leq a \leq0.5$ and $1\leq f \leq10$ to find the best case for $\gamma_{cyclic}$ for each experiment.
Furthermore, we run six times with same seed value and report the mean and unbiased standard derivation.
Other hyperparameter setting is available in the open source code\footnote{https://github.com/yjlee22/CyclicAggregation}.

\subsection{Results}
\subsubsection{Effect of $\gamma_{cyclic}$}\label{sec:effect}
First, in order to check the effect of cyclic server learning rate $\gamma_{cyclic}$, we visualize test accuracy respect to $i_g$ with FedAvg \cite{mcmahan2017communication} and various image classification tasks.
Note that the target performance (dashed line) is the maximum test accuracy obtained by FedAvg with $\gamma_{fixed}$ for each dataset.
According to Fig.~\ref{fig:4}, we can observe that at the beginning of $i_g$, the performance with $\gamma_{cyclic}$ and $\gamma_{fixed}$ tend to be similar.
However, after a certain iteration $i_g$, utilizng cyclic learing rate $\gamma_{cyclic}$ converges faster to the target performance as compared to the coventional fixed rate $\gamma_{fixed}$.

\subsubsection{Communication efficiency}
To investigate the communication cost, we examine the convergence rate of state-of-the-art FL techniques over various image classification tasks with $\gamma_{cyclic}$ and $\gamma_{fixed}$.
Precisely, we examine the minimum number of iteration $i_g$ to reach the target performance $T$. Note that we utilize grid search to find the best cases of $\gamma_{cyclic}$ as in the previous experiment.
According to Table. \ref{table:1}, it shows that $\gamma_{cyclic}$ requires less number of $i_g$ to achieve the target performance $T$ of each dataset than $\gamma_{fixed}$ for every FL techniques.
Specifically, we observe that convergence rate is more than doubled as compared to FedAvg with FMNIST dataset.
In practice, since the coomunication cost is the most key factor of FL, the proposed $\gamma_{cyclic}$
is preferable than conventional $\gamma_{fixed}$.

\subsubsection{Performance}
To investigate the maximum performance on the benchmark dataset, we examine the maximum test accuracy over sufficient G with $\gamma_{cyclic}$ and $\gamma_{fixed}$.
According to Table. \ref{table:2}, it is shown that regardless of FL technique and dataset, $\gamma_{cyclic}$ enhances also the performance.

\section{Conclusion}\label{sec:conclusion}
In this paper, we proposed cyclic aggregation rule to alleviate the client-drift problem without any additional computational cost both at the central server side and the client side. The proposed cyclic aggregation rule with cyclic learning rate can be applied to most of existing FL techniques, and we numerically validated its benefit in terms of both convergence speed and generalization performance. Future work may consider hyperparameter optimization technique to find the optimal frequency and amplitude.

\bibliographystyle{IEEEbib}
\bibliography{reference}

\end{document}